%% file: Plausibilization_of_sensor_objects_with_flocking_behaviour.tex
\documentclass[conference, a4paper]{IEEEtran}

\usepackage{amsmath}   
\usepackage{amssymb}
\usepackage{amsfonts}  
\usepackage{graphicx}  
\usepackage{url}       
\usepackage{cite}
\usepackage{citesort}   
\usepackage{balance}
\usepackage{subfig}   
\usepackage{xr}
\usepackage{multirow}
\usepackage{standalone}
\usepackage{pgf}
\usepackage{pgfplots}
\usepackage{verbatim}
\usepackage{tikz}
\usetikzlibrary{arrows.meta,positioning, shapes, patterns, calc, through}
\tikzstyle{vertex}=[circle, draw, inner sep=1pt, minimum size=6pt]
\pgfplotsset{compat=1.17}
\include{Abbr}

\IEEEoverridecommandlockouts 

\begin{document}

\renewcommand{\baselinestretch}{0.96} 

\title{Sensor Object Plausibilization with Boids Flocking Algorithm}

\author{ \authorblockN{Christopher Knievel} \authorblockN{University
    of Applied Sciences HTWG Konstanz}
  \authorblockN{Alfred-Wachtel-Str.~8, 78462 Konstanz, Germany}
  \authorblockA{Email: cknievel@htwg-konstanz.de} \and
  \authorblockN{Lars Krueger} \authorblockN{Continental - A.D.C. GmbH}
  \authorblockN{Lise-Meitner-Str.~10, 89081 Ulm, Germany}
  \authorblockA{Email: Lars.2.Krueger@continental-corporation.de }
  }

\maketitle

\begin{abstract}
  Driver assistance systems are increasingly becoming part of the
  standard equipment of vehicles and thus contribute to road
  safety. However, as they become more widespread, the requirements
  for cost efficiency are also increasing, and so few and inexpensive
  sensors are used in these systems.  Especially in challenging
  situations, this leads to the fact that target discrimination cannot
  be ensured which in turn leads to a false reaction of the driver
  assistance system. Typically, the interaction between moving traffic
  participants is not modeled directly in the environmental model so
  that tracked objects can split, merge or disappear. The Boids
  flocking algorithm is used to model the interaction between road
  users on already tracked objects by applying the movement rules
  (separation, cohesion, alignment) on the boids. This facilitates the
  creation of semantic neighborhood information between road users.
  We show in a comprehensive simulation that with only 7 boids per
  traffic participant, the estimated median separatation between
  objects can improve from 2.4~m to 3~m for a ground truth of
  3.7~m. The bottom percentile improves from 1.85~m to 2.8~m.
\end{abstract}

\IEEEpeerreviewmaketitle

\section{Introduction}
\label{sec:Introduction}

Active safety measures such as Advanced Driver Assistance Systems
(ADAS) are an elementary component of road-safety, especially with
high market penetration~\cite{Lundgren2006}. In order to further
increase road safety, the EU regulation
2019/2144~\cite{RegulationEU2019/21442019} obliges all OEMs to install
an emergency brake assist (EBA) as well as lane departure warning
system. Hence, as vehicles need to be equipped with the necessary
sensors, the main goal is on cost efficiency, utilizing as few sensors
while offering as many functions possible. As a consequence, the
comfort function adaptive cruise control (ACC) becomes a default
function also in the lower segments. Adaptive cruise control and other
SAE-L1 advanced driver assistance systems have been studied
extensively, see also~\cite{Eskandarian2003}.

For ACC, radar-only as well as radar+camera solutions are often used
as sensor configurations. Despite budget sensors, the expectation
especially on comfort function is high. Ambiguities of the radar
sensors due to low angular resolution are predominant and cannot be
resolved completely by the camera. As a consequence, target
association is increasingly hard which leads to so-called ghost
objects and/or lateral position ambiguities~\cite{Folster2005}.
As a result of these ambiguities, false reactions of the system can occur, such as driver-take over scenarios from a field-operation test, which have been
investigated in~\cite{Weinberger2001}. The cause is often not only a
missing detection of the ACC-relevant vehicle but of a false
interpretation of the scenario, i.e. that the detected vehicle has
been falsely assessed to be not relevant, due to, for example, a false
lane association. Target discrimination focuses on the correct lane
association of vehicles to corresponding
lanes~\cite{Zhang2012,Song2019}.

We propose to use the Boids flocking algorithm~\cite{Reynolds1987,Reynolds1999}, which has been introduced by Reynolds to model the motion of bird flocks or fish school, to model the interaction between traffic participants with each other as well as with the environment.

A flocking algorithm has been used in~\cite{Hayashi2016} to derive a
control algorithm for multiple non-holonomic cars. Each car has been
modeled as an individual boid which behaves according to the movement
rules of the typical boid model, which are cohesion, alignment, and
separation. The proposed control managed a collision-free path of all
boids, however, the usecase was limited to a straight highway and,
eventually, the assumption that all cars driving in the same direction
will converge to the same speed does not hold in reality.

In this paper, an individual flock of boids is created to follow each
detected vehicle, i.e.~for $N_v$ vehicles, $N_f$ flocks with $N_b$
boids each are generated. The corresponding detected vehicle acts as a lead
for the swarm to follow. The aforementioned movement rules are applied
to each boid of each flock. In addition, a further rule is introduced
in order to repel the individual flocks to each other in lateral
direction. Thereby ensuring that the flocks maintain a lateral
distance as long as they are within a given longitudinal distance.
Since the boids of a flock are affected not only by the detected lead
vehicle but also by the other boids of its own flock as well as by the
neighboring flocks, the effects of the position uncertainties of an
object on the target discrimination can be mitigated by using the
average lateral distance between flocks.

The remainder of this paper is organized as follows:
Sec.~\ref{sec:obj_plaus_boids} introduces the Boids flocking algorithm
including the proposed extensions in order to facilitate an object
plausibilization. Additionally, the complexity of the proposed
flocking algorithm is analyzed with respect to the
flock size. Simulation results are presented in
Sec.~\ref{sec:numerical_results}. Finally, Sec.~\ref{sec:conclusion}
draws the conclusion.

\section{Object Plausibilization with the Boids Flocking Algorithm}
\label{sec:obj_plaus_boids}

\subsection{Boids Flocking Algorithm}
\label{sec:boid-flock-algor}

\begin{figure}[htbp]
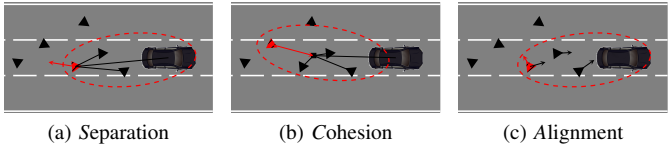

  \centering \subfloat[{\textit Separation}]{
    \includestandalone[width=0.31\linewidth,keepaspectratio]{pic/boid_separation}
  } \subfloat[{\textit Cohesion}]{
    \includestandalone[width=0.31\linewidth,
    keepaspectratio]{pic/boid_cohesion} } \subfloat[{\textit
    Alignment}]{ \includestandalone[width=0.31\linewidth,
    keepaspectratio]{pic/boid_alignment} }
  \caption{Three main rules for boid movement within a flock. The $i$-th boid, for which the rules are calculated for, is given in red color. }
  \label{fig:boid_movement_rules_flock}
\end{figure}

Reynolds~\cite{Reynolds1987} introduced three main rules describing
the movement of boids as an interaction between the individuals of one
flock.  The movement of each boid is influenced by
\begin{itemize}
\item \textbf{Separation}: The tendency of a boid to maintain a
  certain distance from the other boids within the visible range.
\item \textbf{Cohesion}: The tendency of a boid to move to the average
  position of the boids within the visible range.
\item \textbf{Alignment}: The tendency of a boid to align itself with
  the boids within the visible range with respect to orientation and
  velocity.
\end{itemize}
The three basic rules are applied to each boid and each flock within a
certain radius are listed below and depicted in
Figure~\ref{fig:boid_movement_rules_flock}.

We define the boids of the $j$-th flock as
\begin{align}
  \label{eq:boid}
  \setBoidsFlock{j} = \left \lbrace \boid{1,j},\boid{2,j},\ldots \boid{\Nb,j} \right\rbrace
\end{align}
with $j=1,\ldots\Nf$, where $\Nf$ is the number of flocks. Each boid
$\boid{i,j}$ contains the longitudinal and lateral position as well as
velocity. For the sake of readability, we omit the index $j$ of the
flock, as boids do not contain flock-specific information,
i.e. $\boid{i} = [\boidP{i}\, \boidV{i}]^{\mathrm{T}}$, with
$\boidP{i} = [p_{x,i}\, p_{y,i}]^{\mathrm{T}}$,
$\boidV{i} = [v_{x,i}\, v_{y,i}]^{\mathrm{T}}$.

As can be seen in Fig.~\ref{fig:boid_movement_rules_flock}, the
visible range of a boid (field-of-view) is modeled as an ellipse,
opposed to typically a circular section (cf.~\cite{Reynolds1999}), in
order to consider the fact that vehicles are typically driving within
the driving lanes. As a consequence, any boid that deviates too far
laterally from the swarm is no longer considered by the swam. The
deviating boid can nevertheless perceive the swarm and is influenced
by it in its movement.

Therefore, the set of boids, which are within the field-of-view of the
$i$-th boid, is given by
\begin{align}
  \label{eq:set_boids_fov}
  \setBoidsFoV{i} = \left\lbrace \boid{j}  \left| (\boidP{j} - \boidP{i})^{\mathrm{T}} \mathbf{M}^{-1} (\boidP{j} - \boidP{i}) \leq 1 \right. \right\rbrace
\end{align}
with
\begin{align*}
  \mathbf{M}= \!\left[\!\begin{array}{cc} \cos \varphi_i & - \sin \varphi_i \\ \sin \varphi_i & \cos \varphi_i \end{array}\!\right]\! \left[\!\begin{array}{cc} a^2 & 0 \\ 0 & b^2 \end{array}\!\right] \! \left[ \!\begin{array}{cc} \cos \varphi_i & -\sin \varphi_i \\ \sin \varphi_i & \cos \varphi_i \end{array}\!\right]^{\mathrm{T}},
\end{align*}
where the parameter of the ellipse are:
\begin{itemize}
\item $\varphi_i$, the orientation angle of the boid given by
  $\arctan(v_{y,i} / v_{x,i})$; Note that the coordinate system
  according to ISO 8855~\cite{ISO8855_2011} is used, which means that
  the x-coordinate is in longitudinal and the y-coordinate in lateral
  direction.
\item $a$, the length of the first principal axis of the ellipse;
\item $b$, the length of the second principal axis of the ellipse.
\item $|\setBoidsFoV{i}|$, the cardinality of the subset:
  $\Nbfov{i}$. Note that the $i$-th boid is not included in the set
  $\setBoidsFoV{i}$; therefore, $|\setBoidsFoV{i}| < |\setBoids|$.
\end{itemize}

In the following, the main steering rules are described.
\subsection{Movement and interaction rules}
\label{sec:movem-inter-rules}

\textbf{Separation}: Each boid has a tendency to keep a certain
distance from the other boids in the flock, thus, avoiding a
collision. This behavior ensures that the flock is spread both in
longitudinal and lateral direction, effectively, enlarging the
field-of-view of the swarm. As described earlier, we assume that
vehicles travel mainly within the driving lanes and the separation of
the flock should also take this into account in the way that the boids
have a larger separation in the longitudinal direction than in the
lateral direction. However, this is not directly considered in the
separation rule but in the weighting factor of the rule (see also
\eqref{eq:BoidVelUpdate}). Several variations of the separation rule
exists, whereas we follow the implementation of~\cite{Hartman2006}:
\begin{align}
  \label{eq:sep_rule}
  \Rsep = - \sum_{j=1}^{\Nbfov{i}} \boidP{i} - \boidP{j}.
\end{align}
This means, that the position of all boids visible to the $i$-th boid
is subtracted by the position of the $i$-th boid.

\textbf{Cohesion}: Each boid is attracted towards the perceived center
of the flock. This attraction counteracts the separation rule and
causes the boids not to spread throughout the space. Otherwise, the
boids would quickly lose the interaction with each other, due to the
restricted field-of-view. The cohesion force is calculated by
averaging the position of the $\Nbfov{i}$-boids and subtracting the
result from the position of the $i$-th boid:
\begin{align}
  \label{eq:coh_rule}
  \Rcoh = \frac{1}{\Nbfov{i}} \sum_{j=1}^{\Nbfov{i}}\boidP{j} - \boidP{i}.
\end{align}

Next to the attraction of each boid towards its perceived center of
mass of the visible swarm, additional rules have been introduced
in~\cite{Reynolds1999} with the rule \textbf{Leader Following} of
particular interest, as it describes the tendency of a boid to move
closer to a designated \textit{leader} without actually overtaking the
leader. In our proposed approach, each detected vehicle is a natural
designated leader, which are followed by the boids of the
corresponding swarm.  Hence, the attracting force of the leader is
given by
\begin{align}
  \label{eq:coh_leader}
  \Rcohl = \boidP{l} - \boidP{i}.
\end{align}

\textbf{Alignment}: Since every boid of a flock is supposed to follow
the same designated leader, it stands to reason that eventually every
member of the flock should have the same velocity. The alignment rule
is calculated similarly to the cohesion rule, where the average of the
perceived velocity is calculated first and the velocity of the $i$-th
boid is subtracted from it:
\begin{equation}
  \label{eq:ali_rule}
  \Rali = \frac{1}{\Nbfov{i}}  \sum_{j=1}^{\Nbfov{i}}\boidV{j} - \boidV{i}.
\end{equation}

\subsection{Flock repulsion rule}
\label{sec:flock-repulsion-rule}

An interaction between flocks is introduced in this paper, which is
described by the \textbf{flock repulsion} behavior. The idea is that
the repulsive forces of the neighboring flock supports a clear
separation of the boids and thus of the swarms, so that target
discrimination is facilitated even if the object position is
imprecise. The flock repulsion is denoted by $\Rrep$ and is calculated
in two steps.

First, the perceived center of the neighboring flock from the
point-of-view of the $i$-th boid is calculated by
\begin{align}
  \label{eq:center-flock-rep}
  \rrep{i}(:,k) = \frac{1}{\Nbfov{k}} \sum_{j=1}^{\Nbfov{k}} \boidP{j}.
\end{align}
Note here, that multiple flocks, for example on the left and right
lane, can be perceived by one boid. It follows that the perceived
center of mass of the neighboring swarms $\rrep{i}$ are represented in
a matrix of size $[2 \times \Nfx{i}]$, with $\Nfx{i}$ the number of
visible flocks by the $i$-th boid.

\begin{figure}[htbp]
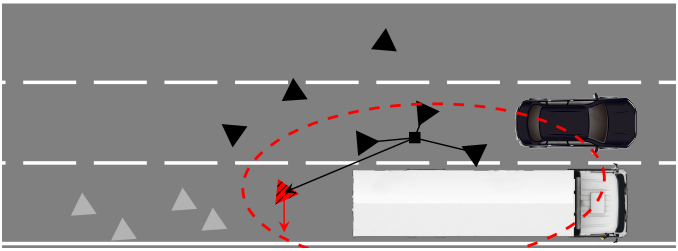

  \centering \includestandalone[width=\linewidth,keepaspectratio]
  {pic/boid_repulsion}
  \caption{Exemplary illustration of the flock repulsion rule for two flocks shown in grey and black color. The red boid, which belongs to the grey flock, uses the average position of the observed black flock.}
  \label{fig:flock-repulsion}
\end{figure}

The repulsing force can then be calculated with an exponential
function whose value is exponentially decreasing with increasing
distance of the swarms, with
\begin{align}
  \label{eq:flock_repulsion}
  \Rrep = \pm \exp\left(\grep - |\boidP{i} - \rrep{i}|\right),
\end{align}
where the sign $\pm$ depends whether the flock is on the left or right
side, respectively.  The value of the factor $\grep$ was chosen so
that the repulsive force is close to zero when the swarms have a
distance of approximately one lane width. Contrary to the other rules,
the repulsing flock rule has a strong effect on the position of the
boids when the distance is small. The rule is exemplary depicted in
Fig.~\ref{fig:flock-repulsion}.
Although the rule is formulated generally in both longitudinal and
lateral direction, the separation of flocks is only carried out in
lateral direction and is considered in the weighting factor $\wrep$,
which is explained in the following.

\subsection{Position Update}
\label{sec:position-update}

The presented five behavioral rules are combined in an updated
velocity vector $\boidVupdate{i}$ of the $i$-th boid and added to the
velocity vector $\boidV{i}$ of the previous cycle:
\begin{align}
  \label{eq:BoidVelUpdate}
  \boidVupdate{i} = & \boidV{i} + \wcoh \Rcoh + \wcohl \Rcohl \nonumber \\
                    & +  \wali \Rali + \wsep \Rsep + \wrep \Rrep^{\mathrm{T}},
\end{align}
where the weighting factor $\wrep$ for the repulsive behavior is given
by
\begin{align}
  \label{eq:weight-rep-force}
  \wrep = \left[ \begin{array}{cccc} w_{x,1} & w_{x,2} & \ldots & w_{x,\Nfx{i}} \\
                   w_{y,1} & w_{y,2} & \ldots & w_{y,\Nfx{i}}
                 \end{array} \right]
\end{align}
with $w_{x,:} := 0$ and
$w_{y,:} := \mathrm{sgn}(p_{y,i} - \rrep{i}(y,:))$.  The remaining
weighting factors are optimized heuristically and the chosen values
can be found in Table~\ref{tab:para_usecase}.

Given the updated velocity vector, the new position of the $i$-th boid
can be calculated straightforward by
\begin{align}
  \label{eq:boid-pos-update}
  \boidPupdate{i} = \boidP{i} + \boidVupdate{i}.
\end{align}

\subsection{Life-cycle of Boids}
\label{sec:spawning-boids}

Unlike other publications using Boids flocking algorithm, it is
assumed in this paper, that boids have a rather short lifetime,
meaning a boid is spawned and will eventually cease to exist within a
duration of a few hundred update cycles, whereas each cycle is assumed to have a
duration of about $80$~ms.

As soon as a lead vehicle is consistently tracked, boids will be
spawned by the lead vehicle every $100$~ms until $\Nb$ boids per flock
exist. The position of the lead-vehicle as well as the lateral and
longitudinal velocity from the previous cycle will be provided to the
new boid and serve as initial values.

\subsection{Reachability analysis with Dubins path}
\label{sec:reach-analys-dubins}

Simulating the movement of the boids of each flock according
to~\eqref{eq:BoidVelUpdate} and ~\eqref{eq:boid-pos-update}, it can be
seen that the resulting path of each boid is only $G0$-continuous and
that boids may ``jump" sideways. In order to constrain the movement of
boids and keeping in mind that boids shall follow vehicles with
non-holonomic constraints, the reachability of the updated position of
a boid is checked by a path generated using a Dubins
path~\cite{Dubins1957}. Dubins paths consist only of straight paths
(`S') and curve segments with a restricted radius, i.e. left curve
(`L') and right curves (`R'), respectively.

\begin{figure}[htbp]
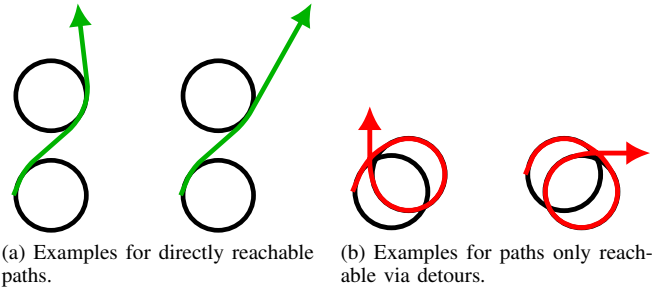

  \centering \subfloat[Examples for directly reachable paths.]{
    \includestandalone[width=0.45\linewidth,
    keepaspectratio]{pic/dubins_path}} \quad \subfloat[Examples for
  paths only reachable via detours.]{
    \includestandalone[width=0.45\linewidth,
    keepaspectratio]{pic/dubins_path_nok}}
  \caption{Exemplary illustration of resulting Dubins paths.}
  \label{fig:dubins_path_eval}
\end{figure}

The minimum radius $\rmin$ is given by the longitudinal velocity $v$
and a fixed maximum lateral acceleration $\alatmax$:
\begin{align*}
  \rmin = \frac{v^2}{\alatmax}.
\end{align*}
The maximum lateral acceleration for each boid is chosen to be
$9~\siAms$, which results in a radius of about $60$~m at a velocity of
$23$~m/s.  In order to calculate a Dubins path from the current
position to the updated target position of a boid, the start and
target pose need to be determined, where as the orientation angles of
start $\varphi_i$ and target pose ($\varphi_i^\prime$) of a boid are
calculated by
\begin{align}
  \label{eq:boidOrientation}
  \varphi_i = \atantwo\left(\frac{\boidV{y,i}}{\boidV{x,i}}\right),\; \varphi_i^\prime = \atantwo\left(\frac{\boidVupdate{y,i}}{\boidVupdate{x,i}}\right)
\end{align}
which yields
$\boidPose{i} = [p_{x,i}, p_{y,i}, \varphi_i]^{\mathrm{T}}$. With the
given start and target pose as well as maximum radius, the resulting
Dubins path will be evaluated. Exemplary evaluations are depicted in
Figure~\ref{fig:dubins_path_eval}, whereas valid paths are given in
green and invalid paths in red.
It can be seen that as soon as detours are required to reach the
target pose, this pose is discarded either due to its position and/or
orientation. Instead, in an iterative process, the target pose is
changed in both position as well as orientation within a small radius
--- and thus speed in lateral and longitudinal direction --- until the
target pose can be reached without detours or until a maximum number
of iterations is reached, which can be used to reduce false reactions of driver assistance systems.




\section{Numerical Results}
\label{sec:numerical_results}

\begin{figure}[htbp]
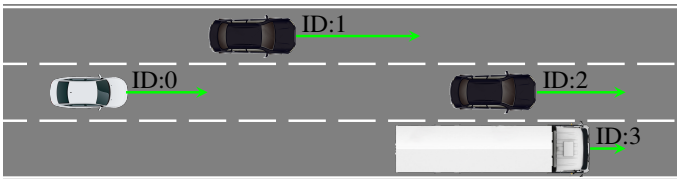

  \centering \includestandalone[width=\linewidth,
  keepaspectratio]{pic/sim_setup}
  \caption{Usecase description}
  \label{fig:usecase}
\end{figure}

A three-lane highway scenario with three target vehicles is considered
in the following with one vehicle driving in each lane and shown in
Figure~\ref{fig:usecase}. The course includes gentle curves as well as
straight sections. The ego vehicle and the preceding vehicle in the
same lane (ID:2) are driving with the same velocity of
$v_{\mathrm{ego}} = v_2 = 25$~m/s at a distance of about $30$~m, which
results in a timegap of $T_\mathrm{G} = d_2 / v_2 = 1.2$~s; a typical
headway distance of an ACC system. A slightly slower vehicle (ID:3) is
driving on the right lane while a faster vehicle (ID:1) is approaching
the ego vehicle from behind, eventually overtaking the ego vehicle and
the other two vehicles.
The parameters of the simulation setup are summarized in
Table~\ref{tab:para_usecase}.  A standard sensor setup for advanced
driver assistance systems is chosen for the ego vehicle, comprising a
long range radar as well as a monocular camera. The detection of each
sensor are subsequently fused providing qualified tracked
objects. These tracked objects serve as potential leaders to create
the individual flock of boids.

As the chosen velocities of the target vehicles (ID:2) and (ID:3)
differ by only $2$~m/s, the reflections of the radar sensors of the
two vehicles are ambiguous and false associations are likely. As a
consequence, the lateral position of a tracked object may be shifted
in lateral direction or worse, reflections of two vehicles are merged
into one tracked object.

\begin{figure}[tp]
  \centering \subfloat[$N_f=3$, $N_b=3$]{
    \includegraphics[width=0.479\linewidth,
    keepaspectratio]{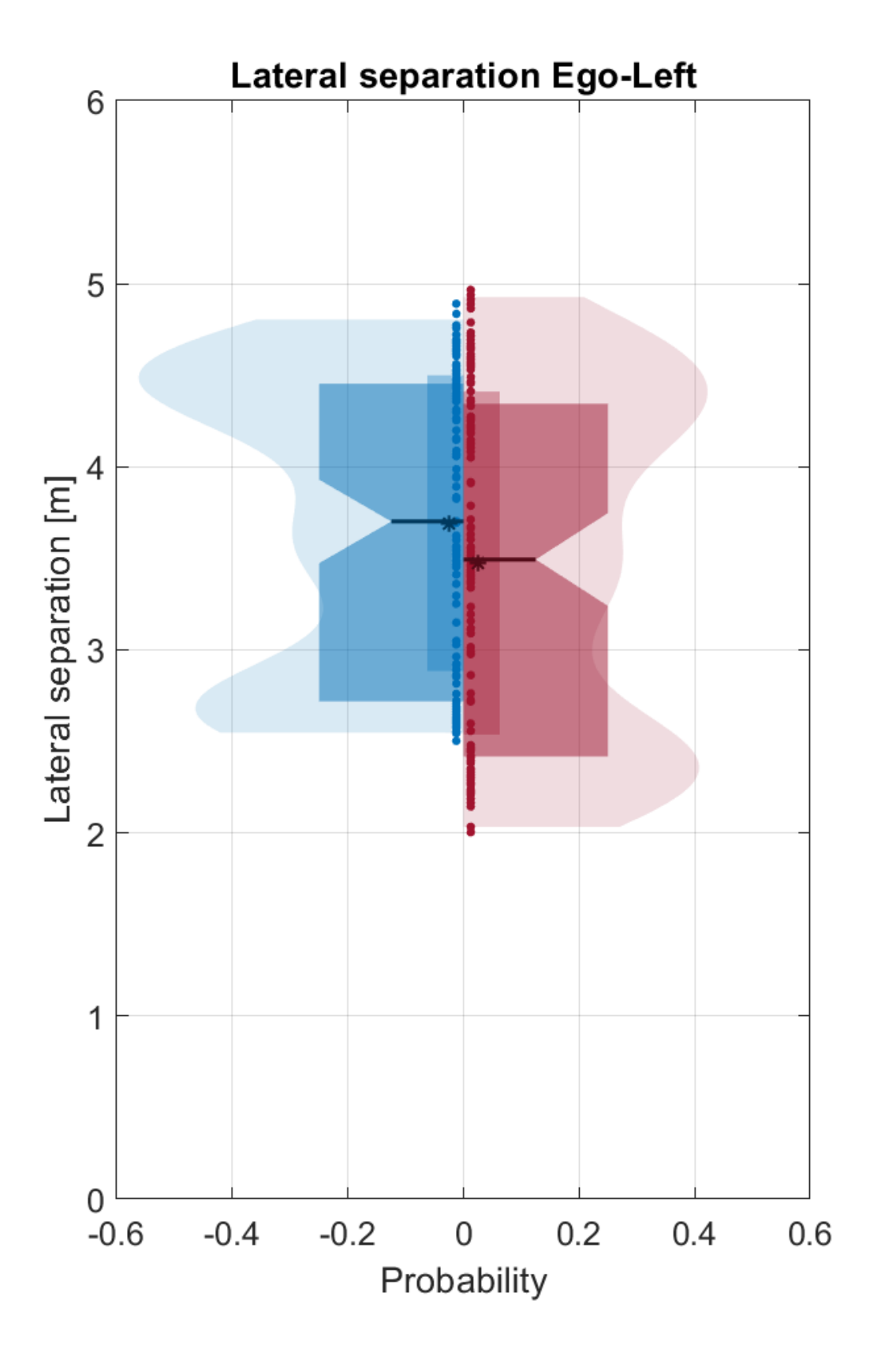}
    \includegraphics[width=0.479\linewidth,
    keepaspectratio]{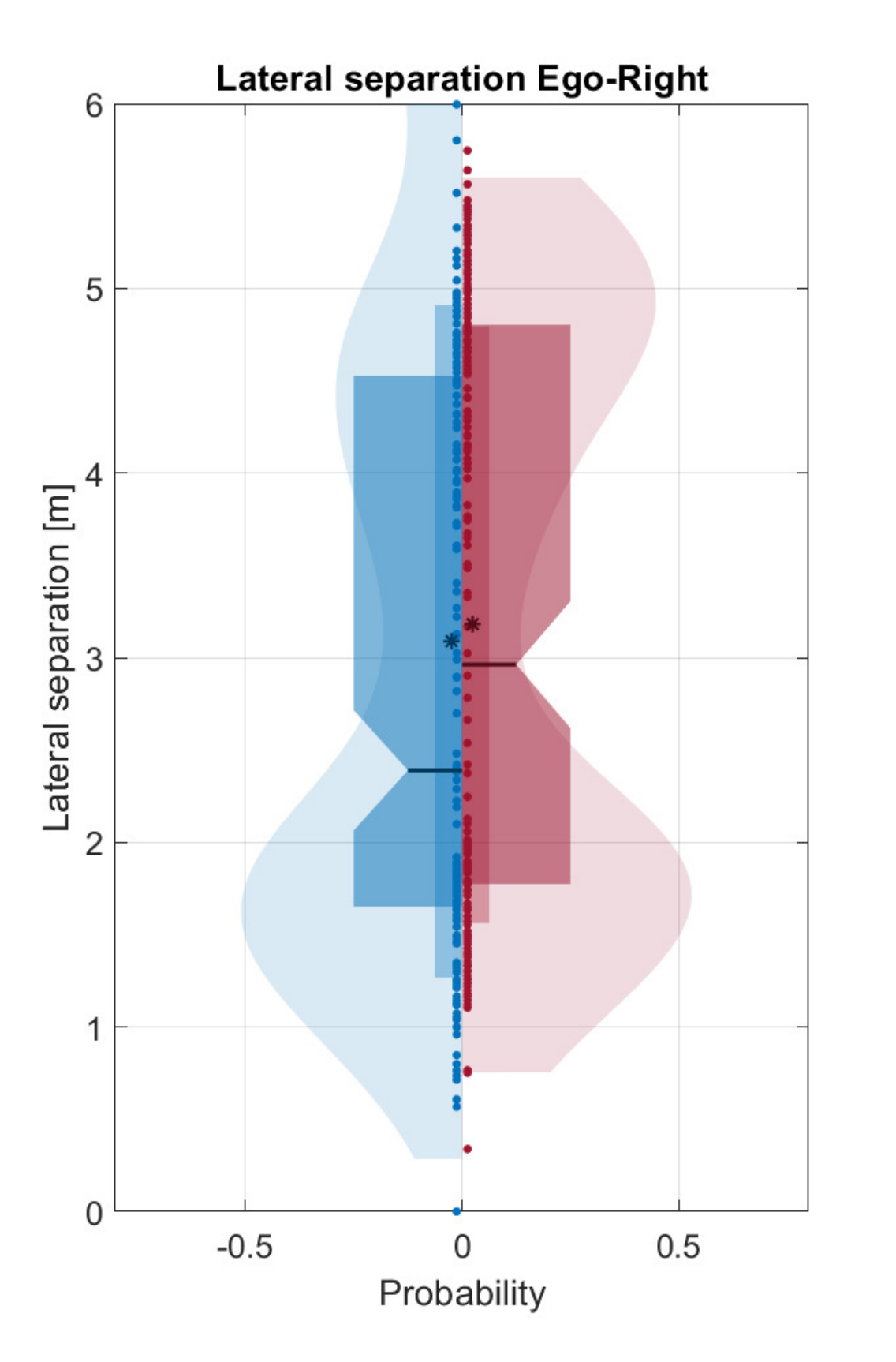}
    \label{fig:violon_a}}
  \\
  \subfloat[$N_f=3$, $N_b=7$]{ \includegraphics[width=0.479\linewidth,
    keepaspectratio]{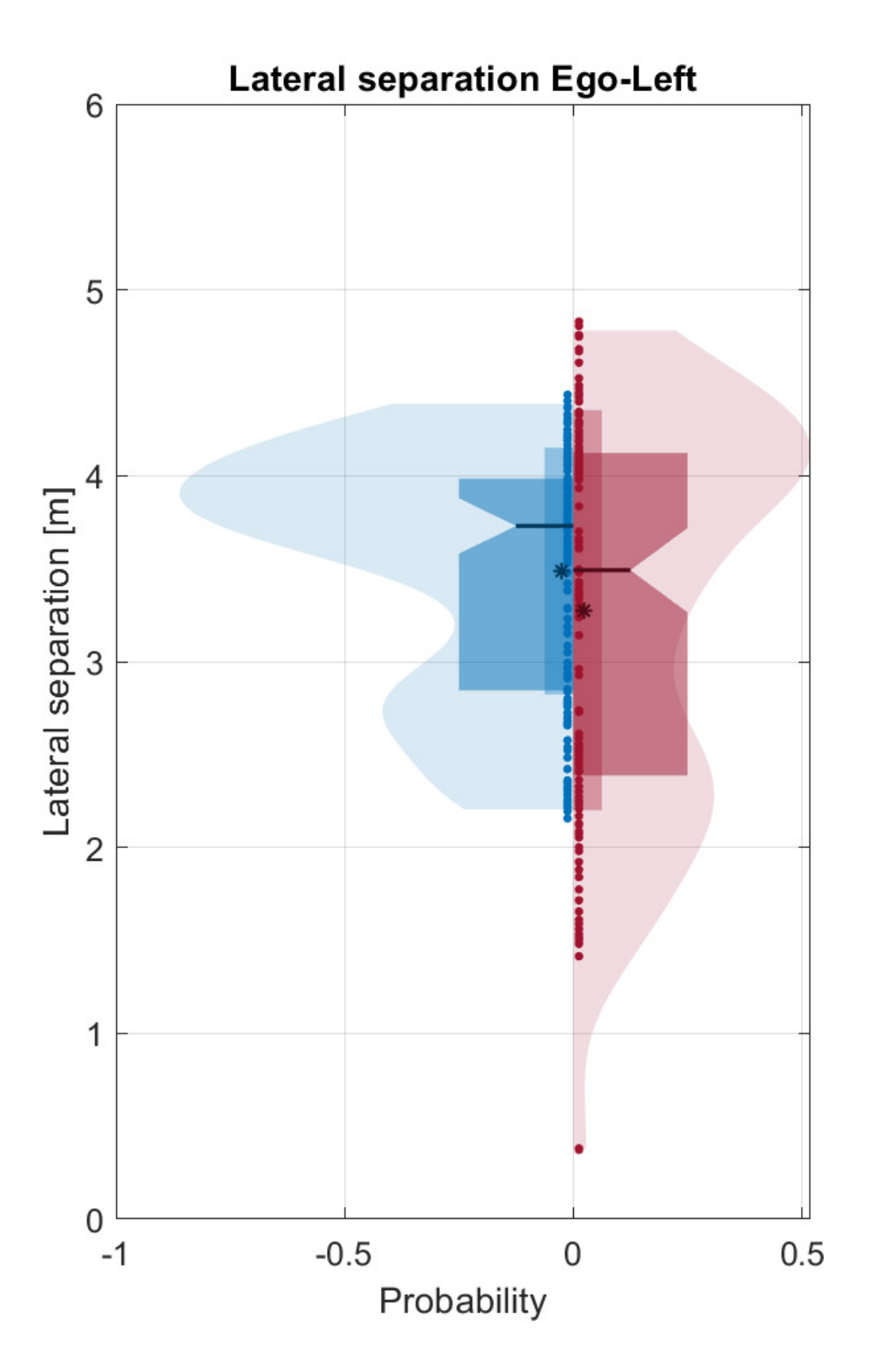}
    \includegraphics[width=0.479\linewidth,
    keepaspectratio]{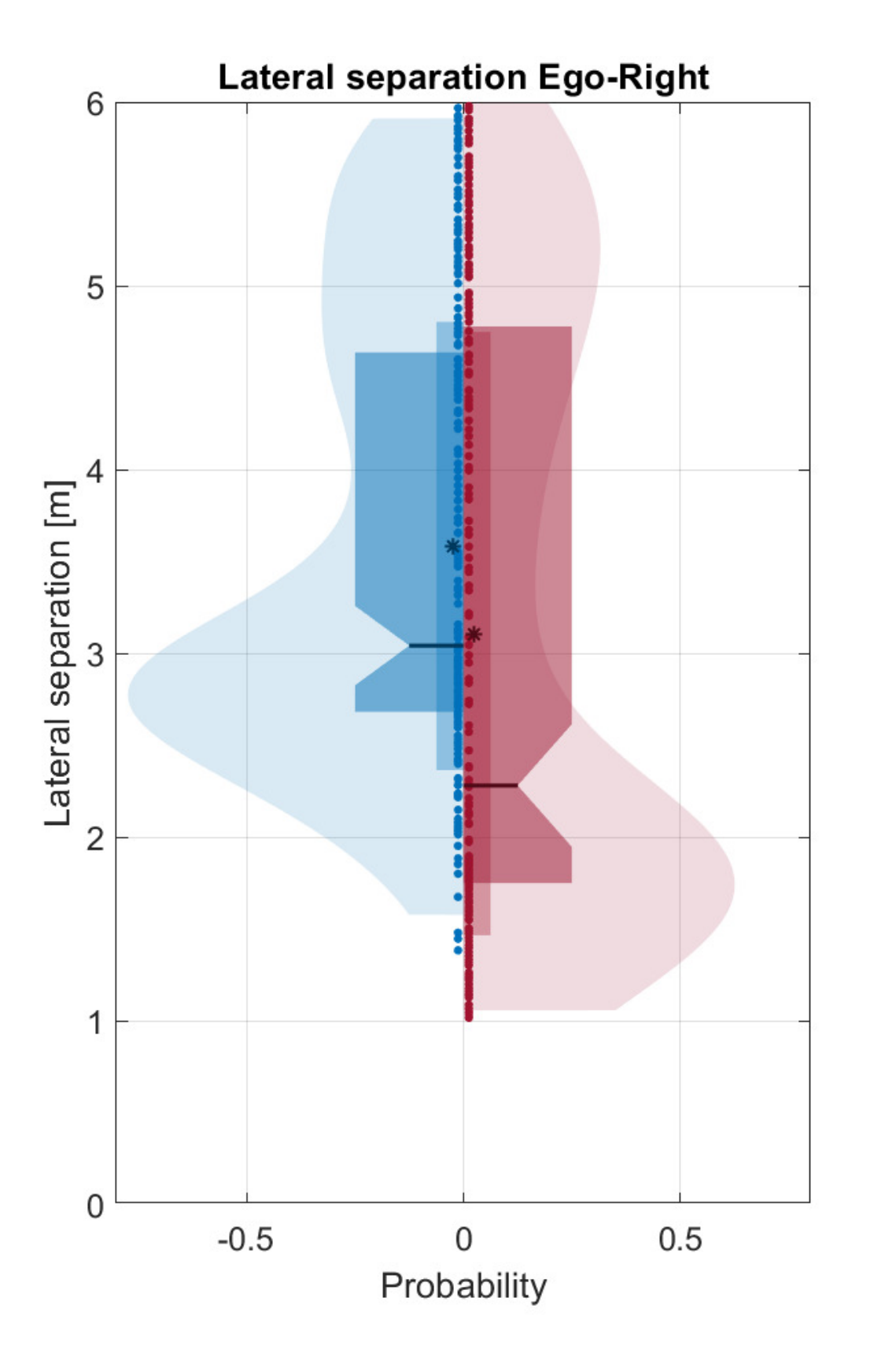}
    \label{fig:violin_b}}
  \\
  \subfloat[$N_f=3$, $N_b=14$]{ \includegraphics[width=0.479\linewidth,
    keepaspectratio]{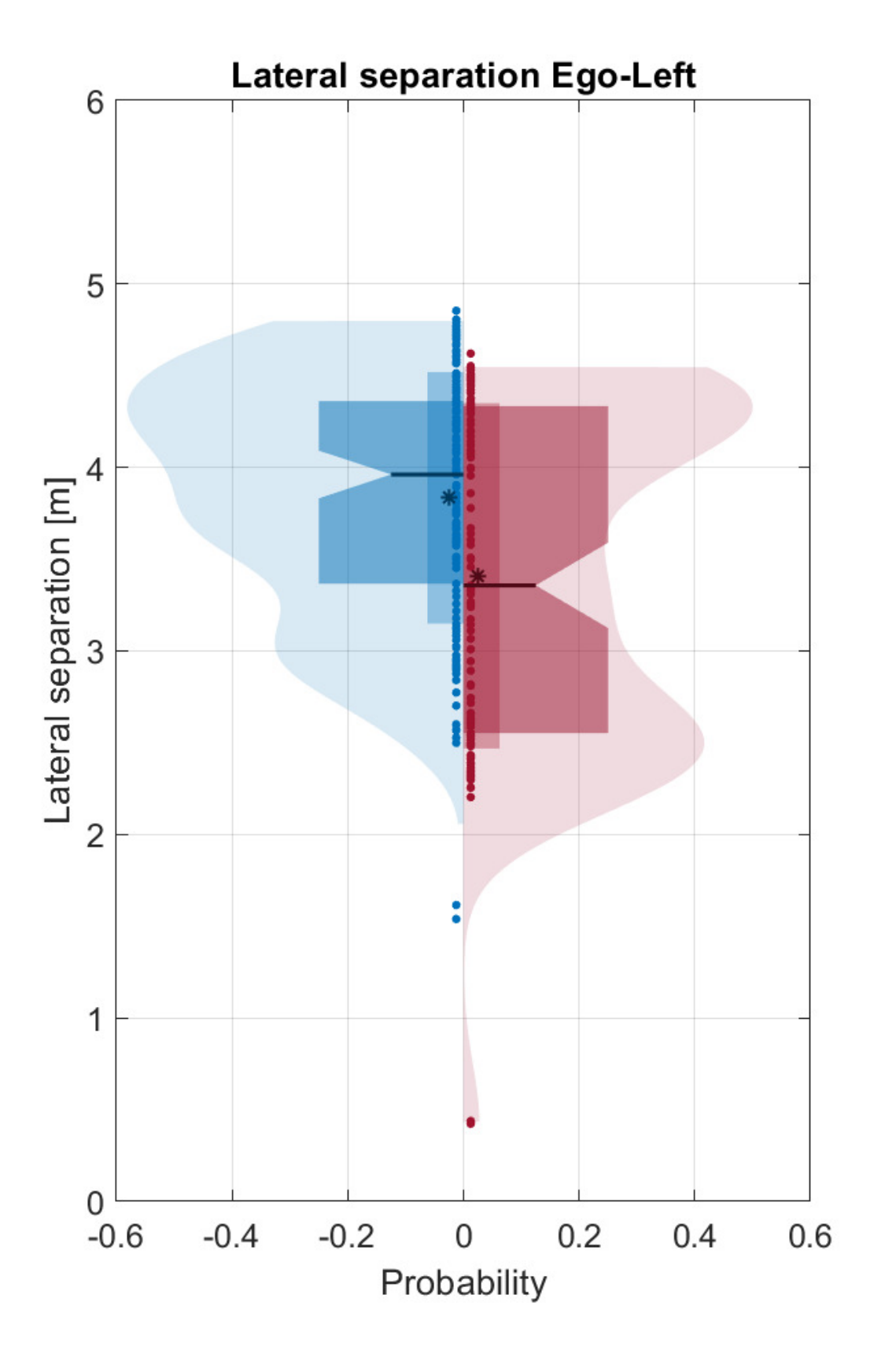}
    \includegraphics[width=0.479\linewidth,
    keepaspectratio]{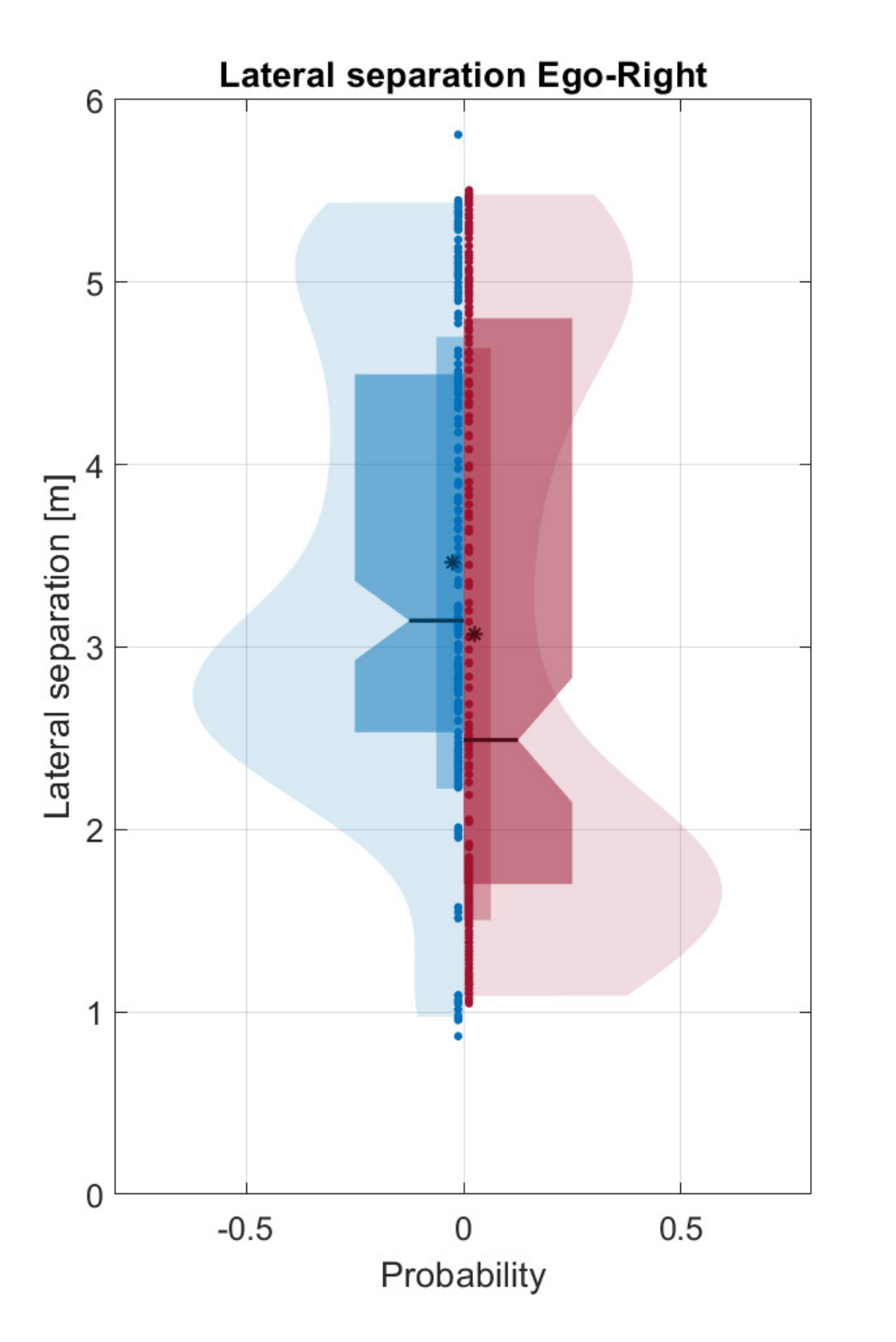}
    \label{fig:violin_c}}
  \caption{\it Split-violin plots for the distribution of the lateral
    distance between vehicles on the ego and left lane (Ego-Left)
    as well as vehicles on the ego and right lane (Ego-Right) .
    Lateral distance of boids in blue (left violin), lateral distance of
    tracked objects (i.e. inputs) in red (right violin).}
  \label{fig:violin_sim_plots}
\end{figure}
The intention of the boids is not to determine a better estimate of
the ground truth position compared to the tracked objects but to
mitigate the shortcomings of the sensors by providing additional
information about the relative position of the vehicles.  Hence, the
relative distance between the vehicles is taken as a measure for the
target discrimination. The distance will be taken from the two pairs
of target vehicles, whereby `Ego-Left' refers to the target
combination (ID:1 and ID:2) and `Ego-Right' to the target combination
(ID:2 and ID:3).

The numerical results are shown in Figure~\ref{fig:violin_sim_plots}
for increasing swarm sizes, combining a boxplot with the probability
distribution, which allows for a better comparison of the two setups.
For the visualization of the numerical results, violin
plots~\cite{Hintze1998} are chosen. Each subplot compares the
distribution of the lateral separation determined by the tracked
objects (in red) and determined by boids (in blue). The median of the
distribution is given by the black horizontal in the center of the
notch, whereas the mean value is denoted by the black
star. Correspondingly, first and third quartile are represented by the
borders of the dark colored area, while the light colored region
ranges from the first to the 99th percentile.

As expected, due to the setup of the scenario, the lateral separation
of the right pair `Ego-Right' is worse than that of the pair
`Ego-Left' due to the smaller relative velocity and therefore increased difficulty for the target discrimination in the environmental model.  The smallest swarm size, with $\Nb=3$
(cf. Fig.~\ref{fig:violon_a}), shows a performance that is inferior to
that of the tracked objects. This becomes clear by the smaller median
of the distribution (2.4~m compared to 3~m) as well as stronger outliers
for the Ego-Right pair.

With increasing swarm size, e.g. $\Nb=7$
(cf. Fig.~\ref{fig:violin_b}), the median of the lateral separation
increases for the boids slightly above 3~m for the Ego-Right pair which
approaches the true lateral separation of 3.7~m. The medians for the
Ego-Left pair for both boids as well as tracked objects are
comparatively close together, which was expected since the target
vehicles (ID:1 to 3) drive parallel to each other only for a short
time due to the higher relative velocity. However, the outliers for
the tracked objects below a lateral separation of 2~m could be
mitigated.

Interestingly, the results are not exclusively improving with a
further increasing swarm size, see Fig.~\ref{fig:violin_c} for
$\Nb=14$. The \textit{separation rule} forces the boids to split along the longitudinal axis, which leads to an increased distance between first and last boid and thus a decreased influence on the average swarm position due the limited field-of-view of each boid.
\begin{table}[t]
  \centering
  \begin{tabular}{|ll|} \hline Velocity ego vehicle ID:0 &
    $v_\mathrm{ego} = 25$~m/s \\ \hline Velocity black car ID:1 &
    $v_1 = 33$~m/s \\ \hline Velocity black car ID:2 & $v_2 = 25$~m/s
    \\\hline Velocity truck ID:3 & $v_3 = 23$~m/s\\ \hline Lateral
    separation ID:1-ID:2 & $d_{12} = 3.2$~m \\ \hline Lateral
    separation ID:2-ID:3 & $d_{23} = 3.7$~m \\ \hline \hline $\wsep$ &
    $[0.15, 0.07]^{\mathrm{T}}$ \\ \hline $\wcoh$ &
    $[0.4, 0.4]^{\mathrm{T}}$ \\ \hline $\wcohl$
    &$[0.4, 0.2]^{\mathrm{T}}$ \\ \hline $\wali$ &
    $[0.3, 0.3]^{\mathrm{T}}$ \\ \hline $\grep$ & 1.5 \\ \hline
  \end{tabular}
  \caption{\it Parameters of the considered usecase as well as for the
    flocking algorithm.}
  \label{tab:para_usecase}
\end{table}

\section{Conclusions}
\label{sec:conclusion}
The Boids flocking algorithm has been evaluated for the target
discrimination in driver assistance systems. By creating an individual
flock for each detected vehicle, together with the presented movement
rules for boids, simulation results can illustrate that the separation of individual vehicles is consistently improved although the tracked objects were partly less than 0.5m apart.

The average lateral position information of a swarm can be used either for lane association or for the improvement of a lane change detection in low-cost driver assistance systems.

So far, only moving traffic participants have been used to create new flocks of boids. In the future, also static infrastructure, such as guard rails, lane
markings, etc. shall be used to create flocks of boids. In combination
with the flock repulsion rule, it will be investigated whether the target separation of parallel driving vehicles can be improved even further.
%
%
\bibliographystyle{IEEEtran} \bibliography{boid}

\end{document}

%% file: Abbr.tex
\newcommand{\Nf}{N_\mathrm{f}}
\newcommand{\Nfx}[1]{N_\mathrm{f,#1}}
\newcommand{\Nb}{N_\mathrm{b}}
\newcommand{\Nbfov}[1]{N_\mathrm{b, FoV_{#1}}}
\newcommand{\boid}[1]{\mathbf{b}_{#1}}
\newcommand{\boidP}[1]{\mathbf{p}_{#1}}
\newcommand{\boidPupdate}[1]{\mathbf{p}^\prime_{#1}}
\newcommand{\boidV}[1]{\mathbf{v}_{#1}}
\newcommand{\boidVupdate}[1]{\mathbf{v}^\prime_{#1}}

\newcommand{\boidPose}[1]{\mathbf{\xi}_{#1}}

\newcommand{\setBoids}{\mathcal{B}}
\newcommand{\setBoidsFlock}[1]{\mathcal{B}_{#1}}
\newcommand{\setBoidsFoV}[1]{\mathcal{B}_{\mathrm{FoV},#1}}

\newcommand{\Rali}{\mathbf{f}_{\mathrm{ali}}}
\newcommand{\Rcoh}{\mathbf{f}_{\mathrm{coh}}}
\newcommand{\Rcohl}{\mathbf{f}_{\mathrm{coh,l}}}
\newcommand{\Rsep}{\mathbf{f}_{\mathrm{sep}}}
\newcommand{\Rrep}{\mathbf{F}_{\mathrm{rep}}}
\newcommand{\grep}{\mathbf{\gamma}_{\mathrm{rep}}}
\newcommand{\rrep}[1]{\mathbf{R}_{\mathrm{rep,#1}}}
\newcommand{\wali}{\mathbf{w}_{\mathrm{ali}}}
\newcommand{\wcoh}{\mathbf{w}_{\mathrm{coh}}}
\newcommand{\wcohl}{\mathbf{w}_{\mathrm{coh,l}}}
\newcommand{\wsep}{\mathbf{w}_{\mathrm{sep}}}
\newcommand{\wrep}{\mathbf{W}_{\mathrm{rep}}}

\newcommand{\alatmax}{a_{\mathrm{lat, max}}}
\newcommand{\rmin}{r_{\mathrm{min}}}

\newcommand{\atantwo}{\mathrm{atan2}}
\newcommand{\siAms}{\mathrm{m}/\mathrm{s}^2}


%% file: pic/boid_separation.tex
\begin{tikzpicture}[
    bg/.style={
        rectangle,
        fill=white,
        rounded corners,
      },
       runway node/.style={
        gray,
        line width=72pt,
        postaction={
          draw,
          white,
          line width=70pt,
        },
        postaction={
          draw,
          gray,
          line width=68pt,
          },
        postaction={
               draw,
               white,
               line width=24pt,
               dash pattern=on 15pt off 5pt, 
             },
             postaction={
               draw,
               gray,
               line width=22pt,
               },
          },
         stopline node/.style={
            white,
            line width=1pt,
          },
     ]

 \def\bcar(#1)#2{%
  \begin{scope}[shift={(#1)}]
    \node{\includegraphics[angle={#2},scale=0.04]{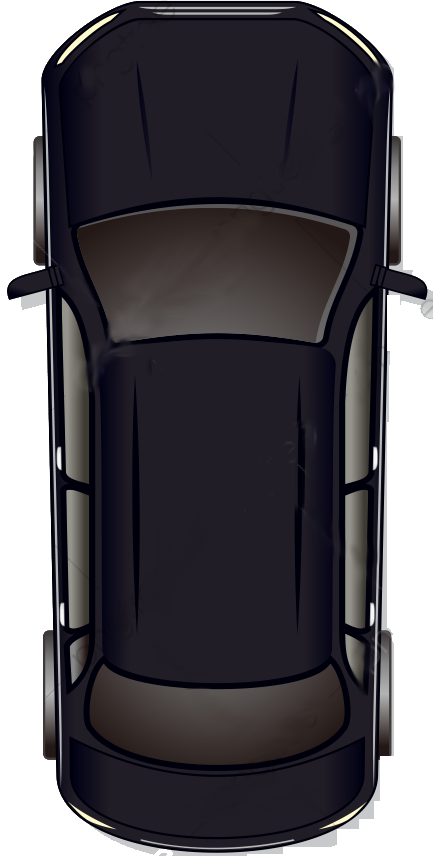}};
  \end{scope}
}


\begin{scope}
  \clip(-1.7,-0.25) rectangle (3,2.25);
\node [bg] at (-1.7,1) {} ;

\draw[runway node] (-2.2,1) -- (3,1);

\bcar(2,1) {270};


\node[fill=black,regular polygon, regular polygon sides=3,inner sep=1.5pt,rotate=65] at (1,0.7) (b1) {};

\node[fill=black,regular polygon, regular polygon sides=3,inner sep=1.5pt,rotate=35] at (0.5,1.1) (b2) {};

\node[fill=red,regular polygon, regular polygon sides=3,inner sep=1.5pt,rotate=35, postaction={pattern=north east lines}] at (-0.1,0.8) (b3) {};

\node[fill=black,regular polygon, regular polygon sides=3,inner sep=1.5pt,rotate=4] at (-0.8,1.3) (b4) {};

\node[fill=black,regular polygon, regular polygon sides=3,inner sep=1.5pt,rotate=55] at (-1.4,0.9) (b5) {};

\node[fill=black,regular polygon, regular polygon sides=3,inner sep=1.5pt,rotate=-5] at (0.1,1.8) (b6) {};

\draw[rotate=5, red, dashed, thick] (1.2,0.85) ellipse (1.5cm and 0.6cm);

\draw (b3.center) -- (b1.center);
\draw (b3.center) -- (b2.center);
\draw (2,1) -- (b3.center);
\draw [-stealth,red] (b3.center) -- (-0.7,0.9);
\end{scope}

\end{tikzpicture}

%% file: pic/boid_cohesion.tex
\begin{tikzpicture}[
    bg/.style={
        rectangle,
        fill=white,
        rounded corners,
      },
       runway node/.style={
        gray,
        line width=72pt,
        postaction={
          draw,
          white,
          line width=70pt,
        },
        postaction={
          draw,
          gray,
          line width=68pt,
          },
        postaction={
               draw,
               white,
               line width=24pt,
               dash pattern=on 15pt off 5pt, 
             },
             postaction={
               draw,
               gray,
               line width=22pt,
               },
          },
         stopline node/.style={
            white,
            line width=1pt,
          },
     ]

 \def\wcar(#1)#2{%
   \begin{scope}[shift={(#1)}]
    \node{\includegraphics[angle={#2},scale=0.04]{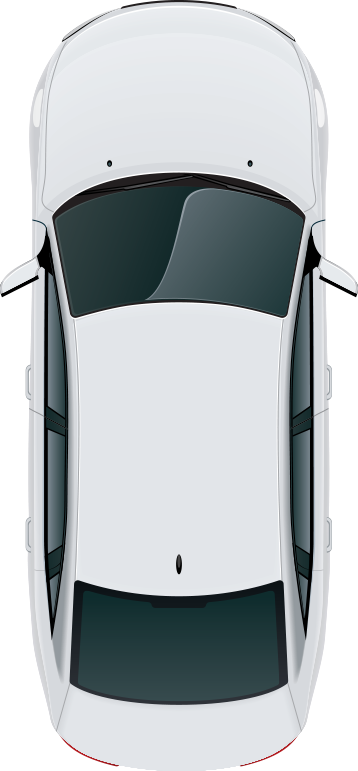}};
  \end{scope}
}

 \def\bcar(#1)#2{%
  \begin{scope}[shift={(#1)}]
    \node{\includegraphics[angle={#2},scale=0.04]{pic/black_car_top.png}};
  \end{scope}
}

 \def\wtruck(#1)#2{%
   \begin{scope}[shift={(#1)}]
    \node{\includegraphics[angle={#2},scale=0.35]{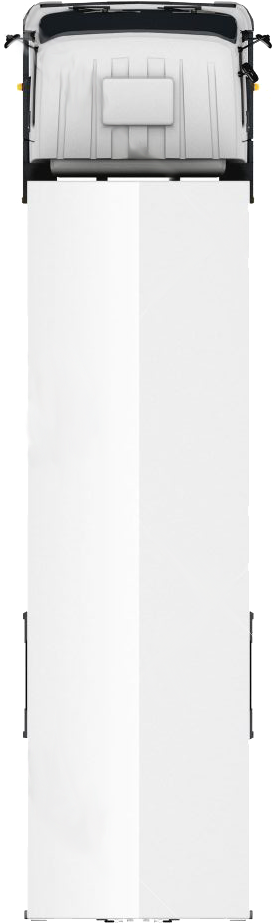}};
  \end{scope}
}

\begin{scope}
  \clip(-1.7,-0.25) rectangle (3,2.25);
\node [bg] at (-1.7,1) {} ;

\draw[runway node] (-2.2,1) -- (3,1);

\bcar(2,1) {270};

\draw[fill=black] (0.1,1) rectangle ++(0.1,0.1)node  (cog) {};

\node[fill=black,regular polygon, regular polygon sides=3,inner sep=1.5pt,rotate=65] at (1,0.7) (b1) {};

\node[fill=black,regular polygon, regular polygon sides=3,inner sep=1.5pt,rotate=35] at (0.5,1.1) (b2) {};

\node[fill=black,regular polygon, regular polygon sides=3,inner sep=1.5pt,rotate=35] at (-0.1,0.8) (b3) {};

\node[fill=red,regular polygon, regular polygon sides=3,inner sep=1.5pt,rotate=4, postaction={pattern=north east lines}] at (-0.8,1.3) (b4) {};

\node[fill=black,regular polygon, regular polygon sides=3,inner sep=1.5pt,rotate=55] at (-1.4,0.9) (b5) {};

\node[fill=black,regular polygon, regular polygon sides=3,inner sep=1.5pt,rotate=-5] at (0.1,1.8) (b6) {};

\draw[rotate=-8, red, dashed, thick] (0.2,1.15) ellipse (1.5cm and 0.6cm);

\draw[-stealth,red] (b4.center) -- ([xshift=-1.5pt,yshift=-1.5pt]cog.center);
\draw (b1.center) -- ([xshift=-1.5pt,yshift=-1.5pt]cog.center);
\draw (b2.center) -- ([xshift=-1.5pt,yshift=-1.5pt]cog.center);
\draw (b3.center) -- ([xshift=-1.5pt,yshift=-1.5pt]cog.center);
\draw (2,1) -- ([xshift=-1.5pt,yshift=-1.5pt]cog.center);
\end{scope}

\end{tikzpicture}

%% file: pic/boid_alignment.tex
\begin{tikzpicture}[
    bg/.style={
        rectangle,
        fill=white,
        rounded corners,
      },
       runway node/.style={
        gray,
        line width=72pt,
        postaction={
          draw,
          white,
          line width=70pt,
        },
        postaction={
          draw,
          gray,
          line width=68pt,
          },
        postaction={
               draw,
               white,
               line width=24pt,
               dash pattern=on 15pt off 5pt, 
             },
             postaction={
               draw,
               gray,
               line width=22pt,
               },
          },
         stopline node/.style={
            white,
            line width=1pt,
          },
     ]

 \def\bcar(#1)#2{%
  \begin{scope}[shift={(#1)}]
    \node{\includegraphics[angle={#2},scale=0.04]{pic/black_car_top.png}};
  \end{scope}
}


\begin{scope}
  \clip(-1.7,-0.25) rectangle (3,2.25);
\node [bg] at (-1.7,1) {} ;

\draw[runway node] (-2.2,1) -- (3,1);

\bcar(2,1) {270};


\node[fill=black,regular polygon, regular polygon sides=3,inner sep=1.5pt,rotate=65] at (1,0.7) (b1) {};

\node[fill=black,regular polygon, regular polygon sides=3,inner sep=1.5pt,rotate=35] at (0.5,1.1) (b2) {};

\node[fill=red,regular polygon, regular polygon sides=3,inner sep=1.5pt,rotate=35, postaction={pattern=north east lines}] at (-0.1,0.8) (b3) {};

\node[fill=black,regular polygon, regular polygon sides=3,inner sep=1.5pt,rotate=4] at (-0.8,1.3) (b4) {};

\node[fill=black,regular polygon, regular polygon sides=3,inner sep=1.5pt,rotate=55] at (-1.4,0.9) (b5) {};

\node[fill=black,regular polygon, regular polygon sides=3,inner sep=1.5pt,rotate=-5] at (0.1,1.8) (b6) {};

\draw[rotate=5, red, dashed, thick] (1.2,0.85) ellipse (1.5cm and 0.6cm);

\draw [-stealth] (b3.center) -- ($(b3.center) + (0.35,0.1)$);
\draw [-stealth,red] (b3.center) -- ($(b3.center) + (-0.15,0.25)$);
\draw [-stealth] (b1.center) -- ($(b1.center) + (0.35,0.25)$);
\draw [-stealth] (b2.center) -- ($(b2.center) + (0.35,0.02)$);
\end{scope}

\end{tikzpicture}

%% file: pic/boid_repulsion.tex
\begin{tikzpicture}[
    bg/.style={
        rectangle,
        fill=white,
        rounded corners,
      },
       runway node/.style={
        gray,
        line width=72pt,
        postaction={
          draw,
          white,
          line width=70pt,
        },
        postaction={
          draw,
          gray,
          line width=68pt,
          },
        postaction={
               draw,
               white,
               line width=24pt,
               dash pattern=on 15pt off 5pt, 
             },
             postaction={
               draw,
               gray,
               line width=22pt,
               },
          },
         stopline node/.style={
            white,
            line width=1pt,
          },
     ]

 \def\wcar(#1)#2{%
   \begin{scope}[shift={(#1)}]
    \node{\includegraphics[angle={#2},scale=0.04]{pic/white_car_top.png}};
  \end{scope}
}

 \def\bcar(#1)#2{%
  \begin{scope}[shift={(#1)}]
    \node{\includegraphics[angle={#2},scale=0.04]{pic/black_car_top.png}};
  \end{scope}
}

 \def\wtruck(#1)#2{%
   \begin{scope}[shift={(#1)}]
    \node{\includegraphics[angle={#2},scale=0.35]{pic/white_truck_top.png}};
  \end{scope}
}

\begin{scope}
  \clip(-1.7,-0.25) rectangle (5,2.25);
\node [bg] at (-1.7,1) {} ;

\draw[runway node] (-2.2,1) -- (5,1);

\bcar(4,1) {270};
\wtruck(3.15,0.2){270};

\draw[fill=black] (2.35,0.8) rectangle ++(0.1,0.1)node  (cog) {};

\node[fill=black,regular polygon, regular polygon sides=3,inner sep=1.5pt,rotate=65] at (3,0.7) (b1) {};

\node[fill=black,regular polygon, regular polygon sides=3,inner sep=1.5pt,rotate=35] at (2.5,1.1) (b2) {};

\node[fill=black,regular polygon, regular polygon sides=3,inner sep=1.5pt,rotate=35] at (1.9,0.8) (b3) {};

\node[fill=black,regular polygon, regular polygon sides=3,inner sep=1.5pt,rotate=4] at (1.2,1.3) (b4) {};

\node[fill=black,regular polygon, regular polygon sides=3,inner sep=1.5pt,rotate=55] at (0.6,0.9) (b5) {};

\node[fill=black,regular polygon, regular polygon sides=3,inner sep=1.5pt,rotate=-5] at (2.1,1.8) (b6) {};

\node[fill=red,regular polygon, regular polygon sides=3,inner sep=1.5pt,rotate=24, postaction={pattern=north east lines}] at (1.1,0.3) (bt1) {};

\node[fill=white!70!black,regular polygon, regular polygon sides=3,inner sep=1.5pt,rotate=4] at (0.4,0) (bt2) {};

\node[fill=white!70!black,regular polygon, regular polygon sides=3,inner sep=1.5pt,rotate=4] at (0.1,0.2) (bt3) {};

\node[fill=white!70!black,regular polygon, regular polygon sides=3,inner sep=1.5pt,rotate=4] at (-0.5,-0.1) (bt4) {};

\node[fill=white!70!black,regular polygon, regular polygon sides=3,inner sep=1.5pt,rotate=4] at (-0.9,0.15) (bt5) {};

\draw[rotate=2, red, dashed, thick] (2.5,0.3) ellipse (1.8cm and 0.8cm);

\draw (b1.center) -- ([xshift=-1.5pt,yshift=-1.5pt]cog.center);
\draw (b2.center) -- ([xshift=-1.5pt,yshift=-1.5pt]cog.center);
\draw (b3.center) -- ([xshift=-1.5pt,yshift=-1.5pt]cog.center);
\draw[-stealth] ([xshift=-1.5pt,yshift=-1.5pt]cog.center) -- (bt1.center);
\draw[-stealth,red] (bt1.center) -- ($(bt1.center) + (0,-0.4)$);
\end{scope}

\end{tikzpicture}

%% file: pic/dubins_path.tex
\begin{tikzpicture}

\node[vertex] (c1) at (0,0) {};
\node[vertex] (c2) at (0,0.3) {};
\draw[-{Latex[length=2pt, width=2pt]}, green!70!black, rounded corners=1pt]  (c1.west) -- (c1.north west)  -- (c2.south east) -- (c2.east) -- ($(c2.north east) + (0,0.2)$);

\node[vertex] (c3) at (0.5,0) {};
\node[vertex] (c4) at (0.5,0.3) {};
\draw[-{Latex[length=2pt, width=2pt]}, green!70!black, rounded corners=1pt] (c3.west)-- (c3.north west)  -- (c4.south east)  -- ($(c4.north east) + (0.2,0.2)$);





\end{tikzpicture}

%% file: pic/dubins_path_nok.tex
\begin{tikzpicture}



\node[vertex] (nc1) at (1.5,0) {};
\node[vertex] (nc2) at (1.55,0.05) {};

\draw[-{Latex[length=2pt, width=2pt]},red, rounded corners=1pt] (nc1.west) -- (nc1.north west)  -- (nc2.north west) -- (nc2.north)  -- (nc2.north east) -- (nc2.east)  -- (nc2.south east) -- (nc2.south) -- (nc2.south west) -- (nc2.west) -- ($(nc2.west) + (0,0.2)$);

\node[vertex] (nc3) at (2,0.05) {};
\node[vertex] (nc4) at (2.05,0) {};

\draw[-{Latex[length=2pt, width=2pt]},red, rounded corners=1pt] (nc3.west) -- (nc3.north west)  -- (nc3.north) -- (nc3.north east)  -- (nc4.north east) -- (nc4.east)  -- (nc4.south east) -- (nc4.south) -- (nc4.south west) -- (nc4.west) -- (nc4.north west) -- (nc4.north) -- ($(nc4.north) + (0.2,0)$);

\end{tikzpicture}

%% file: pic/sim_setup.tex
\begin{tikzpicture}[
    bg/.style={
        rectangle,
        fill=white,
        rounded corners,
      },
       runway node/.style={
        gray,
        line width=72pt,
        postaction={
          draw,
          white,
          line width=70pt,
        },
        postaction={
          draw,
          gray,
          line width=68pt,
          },
        postaction={
               draw,
               white,
               line width=24pt,
               dash pattern=on 15pt off 5pt, 
             },
             postaction={
               draw,
               gray,
               line width=22pt,
               },
          },
         stopline node/.style={
            white,
            line width=1pt,
          },
     ]

 \def\wcar(#1)#2{%
   \begin{scope}[shift={(#1)}]
    \node{\includegraphics[angle={#2},scale=0.04]{pic/white_car_top.png}};
  \end{scope}
}

 \def\bcar(#1)#2{%
  \begin{scope}[shift={(#1)}]
    \node{\includegraphics[angle={#2},scale=0.04]{pic/black_car_top.png}};
  \end{scope}
}

 \def\wtruck(#1)#2{%
   \begin{scope}[shift={(#1)}]
    \node{\includegraphics[angle={#2},scale=0.35]{pic/white_truck_top.png}};
  \end{scope}
}
 
\begin{scope}
 \clip(-6.5,-0.25) rectangle (3,2.25);
\node [bg] at (-1.7,1) {} ;

\draw[runway node] (-6.7,1) -- (3.3,1);

\wcar(-5.3,1){270};
\bcar(-3,1.8){270};
\bcar(0.4,1) {270};
\wtruck(0.4,0.2){270};

\node at (-4.9,1) (W1) {};
\node at (-2.5,1.8) (B1) {};
\node at (1.65,0.2) (T1) {};
\node at (0.9,1)(B2){};

\node at (-0.5,1.8)(B12){};
\node at (-3.5,1)(B22){};
\node at (2.4,0.2)(B32){};
\node at (2.4,1)(B42){};

\draw (B1) edge[out=0,in=180,->,>=stealth,line width=1pt,green] (B12) node[xshift=15pt,yshift=5pt] {ID:1};
\draw (W1) edge[out=0,in=180,->,>=stealth,line width=1pt,green] (B22)node[xshift=15pt,yshift=5pt] {ID:0};
\draw (T1) edge[out=0,in=180,->,>=stealth,line width=1pt,green] (B32)node[xshift=15pt,yshift=5pt] {ID:3};
\draw (B2) edge[out=0,in=180,->,>=stealth,line width=1pt,green] (B42) node[xshift=15pt,yshift=5pt] {ID:2};
 \end{scope}
\end{tikzpicture}